\documentclass[letterpaper, 10 pt, conference]{ieeeconf}  

\IEEEoverridecommandlockouts                              
\usepackage{mwe}
\usepackage{graphics} 
\usepackage{subfigure}
\usepackage{graphicx}
\usepackage{bm}
\usepackage{siunitx}
\usepackage{relsize}
\usepackage{mathptmx}
\usepackage{amsmath} 
\usepackage{amssymb}  
\usepackage{booktabs}
\usepackage{siunitx}
\usepackage{multirow}
\usepackage{makecell}
\usepackage{hyperref}
\usepackage[usenames, dvipsnames,table,xcdraw]{xcolor}
\usepackage{pifont}

\usepackage[capitalise]{cleveref}
\usepackage{bm}
\usepackage{mathtools}

\newcommand{\rsec}                                      [1]     {Section~\ref{#1}}
\renewcommand{\cdot}{}

\newcommand{\vect}[1]{\mathbf{#1}}

\title{\LARGE \bf Parallel Transmission Aware Co-Design: Enhancing Manipulator  \\
	Performance Through Actuation-Space Optimization}
\author{Rohit Kumar$^{1*}$, Melya Boukheddimi$^{1}$, Dennis Mronga$^{1}$, Shivesh Kumar$^{1,2}$, and Frank Kirchner$^{1,3}$
\thanks{This work was done in the CoEx project (Grant Number 01IW24008) funded by the German Aerospace Center (DLR) with
	federal funds from the Federal Ministry of Education and Research (BMBF). The authors thank Christoph Stoeffler and Heiner Peters for helpful discussions.}
\thanks{$^{1}$Robotics Innovation Center, DFKI GmbH, 28359 Bremen, Germany.}%
\thanks{$^{2}$Dynamics Division, Department of Mechanics \& Maritime Sciences, Chalmers University of Technology, Gothenburg, Sweden.}%
\thanks{$^{3}$AG Robotik, University of Bremen, 28359 Bremen, Germany.}%
\thanks{{$^{*}$Corresponding author}:
    \href{mailto:r.kumar@dfki.de}{r.kumar@dfki.de}} 
}

\begin{document}

\maketitle
\thispagestyle{empty}
\pagestyle{empty}

\begin{abstract}
In robotics, structural design and behavior optimization have long been considered separate processes, resulting in the development of systems with limited capabilities. 
Recently, co-design methods have gained popularity, where bi-level formulations are used to simultaneously optimize the robot design and behavior for specific tasks. 
However, most implementations assume a serial or tree-type model of the robot, overlooking the fact that many robot platforms incorporate parallel mechanisms.
In this paper, we present a novel co-design approach that explicitly incorporates parallel coupling constraints into the dynamic model of the robot. 
In this framework, an outer optimization loop focuses on the design parameters, in our case the transmission ratios of a parallel belt-driven manipulator, which map the desired torques from the joint space to the actuation space. 
An inner loop performs trajectory optimization in the actuation space, thus exploiting the entire dynamic range of the manipulator.
We compare the proposed method with a conventional co-design approach based on a simplified tree-type model.
By taking advantage of the actuation space representation, our approach leads to a significant increase in dynamic payload capacity compared to the conventional co-design implementation.
\end{abstract}
\section{Introduction}
\label{sec:intro}

Parallel mechanisms are increasingly used in various types of robots due to their superior stiffness, accuracy and payload capacity (see~\cite{KUMAR2020102367} for a survey). 
In humanoid robots and highly dynamic manipulators, parallel mechanisms can achieve proximal actuation to reduce link inertias. 
While many humanoid systems use four-bar mechanisms~\cite{Apgar2018FastOT}, belt transmissions have also become a popular choice~\cite{Chignoli2020,roig2022,christoph-roland}.
A notable example is the MIT humanoid leg, which employs a parallel belt transmission to achieve efficient force transmission while maintaining a compact and lightweight design~\cite{Chignoli2020}. However, such mechanisms introduce complex coupled constraints that challenge conventional modeling and optimization techniques. 
The inherent complexity of mechanical couplings in these mechanisms pose a considerable challenge to system designers, a fact that advocates a simultaneous optimization of robot design and behavior.
This method, commonly referred to as co-design, offers a promising way to fully exploit a robot's potential. 
This is especially true for parallel mechanisms, as they encompass a much larger design space than serial robots.   
\begin{figure}[!t]
	\centering
	\includegraphics[width=0.99\linewidth]{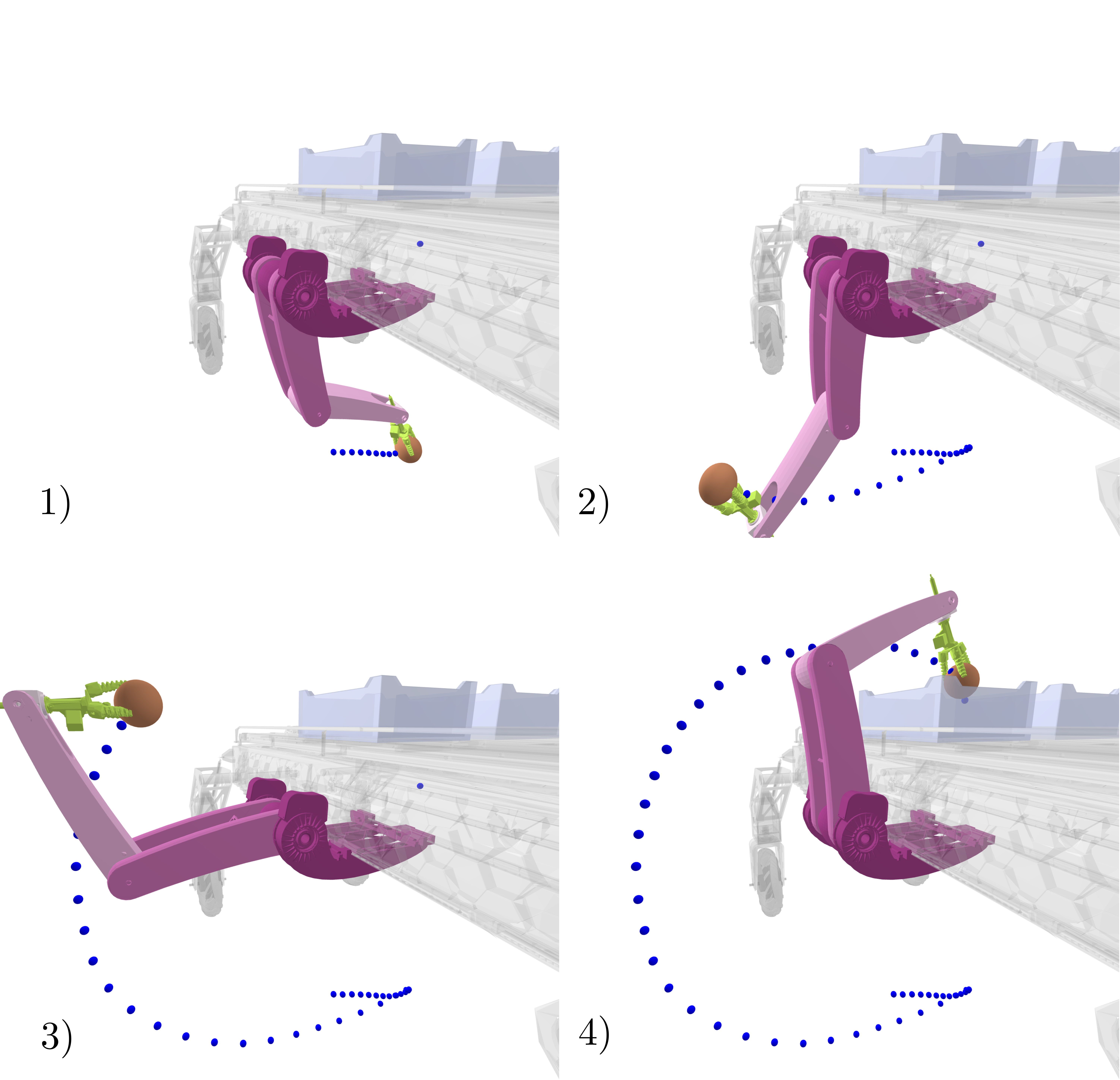}
	\caption{Illustration of the Manipulator lifting 3kg payload (in 1 $\rightarrow$ 4 sequence) through Parallel Transmission Aware Co-Design } 
	\label{fig:roland_arm}
	\vspace{-0.3cm}
\end{figure} 

A large body of work on co-design of robotic systems already exists.
The work presented in~\cite{Cheng2024} shows the co-optimization of a humanoid robot using a combination of reinforcement learning (RL) and evolutionary optimization. 
Similar approaches have been applied to quadrupedal robots~\cite{BelmonteBaeza2022MetaRL,fadini2024,chen2024}, primarily focusing on optimizing link lengths.
Some recent works~\cite{2024icra_girlanda_robust_codesign, 2022iros_lasse_robust_codesign} have studied co-design of simple underactuated systems taking into account the robustness of the controllers.
In contrast, the work in~\cite{Hoffman2025} demonstrates a bi-level optimization of the workspace of a manipulator, including the dimensions of the links and their dynamic properties as well as the choice of actuators. 
In~\cite{Kamadan2017}, the authors present methods for the co-design of robots with variable stiffness actuators. 
\cite{SALUNKHE2022104796} studies the design optimization of a parallel manipulator for surgical applications taking into account only its kinematic properties. 
\cite{disney-paper} presents a computational framework but focuses on link lengths to optimize design and motion trajectories of robotic systems via implicit function theorem.

Common to almost all the aforementioned co-design approaches discussed above is that they can only be applied to serial or tree-type robot architectures. If parallel sub-mechanisms exist in a robot, the given approaches will have to rely on serial abstraction of the robot model, which hides the details of the parallel mechanisms.
Therefore, they cannot exploit the full dynamics and workspace of a robot with parallel mechanisms and ignore the dynamic effects caused by the closed-loop couplings.
This limitation highlights the need for co-design approaches that explicitly account for the dynamic effects and unique constraints of parallel and series-parallel hybrid mechanisms, ensuring a more accurate and efficient optimization process.

\paragraph*{Contribution}
The main contribution of this work is a novel co-design method that considers the overall dynamics of a series-parallel hybrid robot taking into account the effect of its parallel transmissions. 
In contrast to existing approaches that rely on serial or tree-type robot models, our approach performs motion optimization directly in the actuation space of the robot. 
This allows it to fully exploit the available solution space of the robot, resulting in a dynamically more efficient design.
We evaluate our method on a 4-DOF manipulator with parallel belt transmissions~\cite{christoph-roland}, which is supposed to perform a pick-and-place movement as shown in Fig.~\ref{fig:roland_arm}.
The selected design parameters comprise the transmission ratios of the parallel belt couplings. 
The experimental evaluation shows that the resulting robot design offers a significantly higher payload capacity compared to conventional co-design approaches.

\paragraph*{Organization}
This paper is organized as follows:
\rsec{sec:math} provides the mathematical background of the proposed method.
\rsec{sec:methodology} details the proposed co-design methodology.
\rsec{sec:results} presents and discusses the experimental results.
Finally, \rsec{sec:conclusion} concludes the paper and outlines potential directions for future work based on our findings.

\section{Mathematical Preliminaries}
\label{sec:math}
This section presents the mathematical basics required for modeling rigid body systems with and without parallel couplings or kinematic constraints. 

\subsection{Equations of Motion}
For a robotic system with closed-loop or parallel coupling constraints, the equations of motion (EoM) can be expressed in both joint space and actuation space~\cite{kumar2020analytical, rohit-hyrodyn}.

The joint space formulation is:
\begin{equation}
\label{eq:eom_full_implicit}
\vect{H}\ddot{\vect{q}} + \vect{C}(\vect{q},\dot{\vect{q}}) = \vect{\tau}
\end{equation}
where, $\vect{q,\dot{q},\ddot{q}} \in \mathbb{R}^{n}$ are position, velocity, and acceleration of the independent joints, $n$ is the number of independent joints,
$\vect{H(q)} \in \mathbb{R}^{n \times n}$ is the mass-inertia matrix, $\vect{C(q,\dot{q})}\in \mathbb{R}^{n}$ describes the 
Coriolis-centrifugal effects and gravity forces, and $\vect{\tau}\in \mathbb{R}^{n}$ are the joint torques or forces.

The mapping from joint space to actuation space is given by:
\begin{align}
\label{eq:mappings}
	& \vect{q}_u = \gamma(\vect{q}) \\
	& \dot{\vect{q}}_u = \vect{G}  \dot{\vect{q}}  \\
	& \ddot{\vect{q}}_u  =  \vect{G} \ddot{\vect{q}} + \vect{g}_u 
\end{align}
where $\gamma(\vect{q})$ is a function of joint space positions, mapping them to actuation space and $\vect{g}_u = \dot{\vect{G}} \dot{\vect{q}}$. In our case, there is an explicit constraint matrix $\vect{G} \in \mathbb{R}^{n \times m}$ that describes the mapping from joint to actuation space, so $\vect{q}_u = \vect{G}\vect{q} \in \mathbb{R}^m$, where $m$ is the number of actuators. Substituting these equations in (\ref{eq:eom_full_implicit}), we arrive at the actuation space EoM:
\begin{eqnarray}
\vect{G}^{-T}\vect{H}\vect{G}^{-1}\vect{\ddot{q}}_u + \vect{G}^{-T}(\vect{C} - \vect{HG}^{-1}\vect{g}_u) &=& \vect{G}^{-T}\vect{\tau} \\
\label{eq:explicit-eom}
\vect{H}_u\vect{\ddot{q}}_u + \vect{C}_u &=& \vect{\tau}_u
\end{eqnarray}
where $\vect{\tau}_u = \vect{G}^{-T} \vect{\tau} \in \mathbb{R}^m$ are the actuator forces and torques, $\vect{H}_u = \vect{G}^{-T}\vect{HG}^{-1} \in \mathbb{R}^{m \times m}$ is the mass-inertia matrix in actuation space, and $\vect{C}_u = \vect{G}^{-T}(\vect{C} - \vect{HG}^{-1}\vect{g}_u) \in \mathbb{R}^{m \times m}$ contains the Coriolis-centrifugal effects and gravity forces in actuation space.

\subsection{Optimal Control Formulation}
A robot's movement can be formulated as an optimal control (OC) problem, discretized over time. Equation \ref{eq:OCP} defines the optimization problem considered in this work.
\begin{equation}
\begin{aligned}
	\label{eq:OCP}
	\min_{\vect{x}, \vect{u}} \quad & \sum_{k=0}^{N-1} \left( \vect{x}_k^T \vect{Q} \vect{x}_k + \vect{u}_k^T \vect{R} \vect{u}_k  +  \rho_k \|\vect{c}(\vect{x}_k)\|^2 \right)  \\
	\text{s.t.} \quad 
	&  \vect{x}_{k+1} = \vect{f}(\vect{x}_k, \vect{u}_k), \quad k = 0, \dots, N-1 \\
	&  \vect{q}_{\min} \leq \vect{q}_k \leq \vect{q}_{\max}, \quad \text{(Joint limits)} \\
	&  \dot{\vect{q}}_{\min} \leq \dot{\vect{q}}_k \leq \dot{\vect{q}}_{\max}, \quad \text{(Velocity limits)} \\
	&  \vect{u}_{\min} \leq \vect{u}_k \leq \vect{u}_{\max}, \quad \text{(Torque limits)} \\
	&  \vect{x}_0 = \vect{x}_{\text{init}}, \quad \text{(Initial state constraint)} \\
	&  \vect{x}_N = \vect{x}_{\text{final}}, \quad \text{(Final state constraint)}
	\end{aligned}
\end{equation}
where $\vect{x}= [\vect{q,\dot{q}}]^{T} \in \mathbb{R}^{2n}$ corresponds to the robot state, $\vect{Q}^{2n \times 2n}$ and $\vect{R}^{2n \times 2n}$ are diagonal weight matrices associated with the state and control regularization, respectively, $\vect{f}$ represents the discretized dynamics and $\vect{c}(\vect{x})$ imposes Cartesian boundary constraints for all node points as a cost, where $\vect{\rho}$ is the corresponding weighting factor.
In this work, the OC formulation is transcribed into a non-linear programming problem (NLP) using CasADi \cite{casadi} and solved with the IPOPT (Interior Point OPTimizer) solver \cite{ipopt}.
The EoM described in the previous section serve as the basis for formulating the co-design approach in both joint space and actuation space. 
These fundamental concepts are elaborated in detail in the following section.

\section{Methodology}
\label{sec:methodology} 
\begin{figure*}[!htpb]
	\centering
	\includegraphics[width=0.9\linewidth]{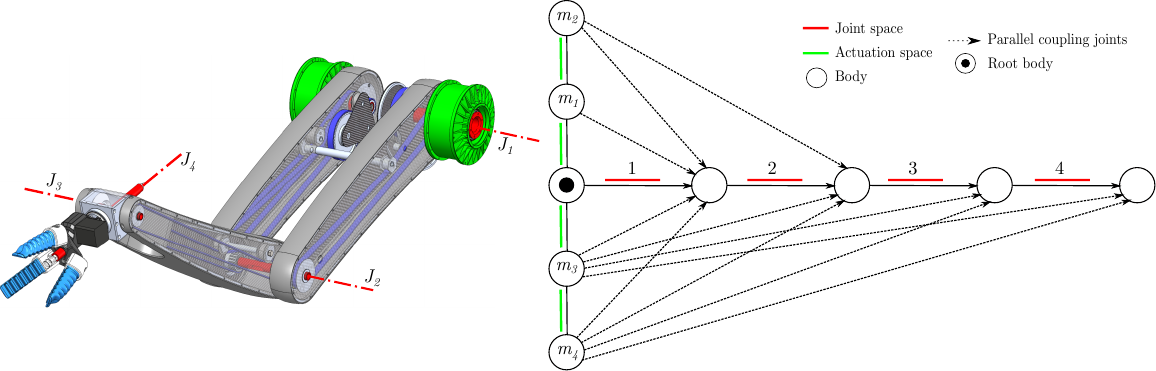}
	\caption{CAD model (left) and topological representation (right) of joint space and actuation space} 
	\label{fig:topological-graph}
	\vspace{-0.3cm}
\end{figure*} 
%

This section presents the coupling constraints for the parallel belt driven robotic manipulator described in \cite{christoph-roland}. 
The manipulator  and its topological graph, shown in Fig.~\ref{fig:topological-graph}, is designed for high-speed pick-and-place operations in automated fruit harvesting. 
To minimize moving inertia and enhance dynamic performance, all actuators are base-mounted, utilizing parallel coupling mechanisms for joint actuation.
The total weight of the manipulator is approximately 5.3 kg, with moving components accounting for 1.6 kg. 
This lightweight design ensures energy efficiency but limits the payload capacity to fruits weighing $< 1$kg.

The limited payload capacity restricts the applicability of the robotic arm to a narrow subset of agricultural tasks, making it unsuitable for harvesting larger fruits. 
To enhance the payload capacity of the manipulator and broaden its application range, we propose a novel co-design approach that incorporates the parallel coupling constraints into the optimization process for design and behavior in a bi-level manner. 
The goal is to find design parameters that allow the manipulator to pick fruits in the range of 1 to 3 kg. 
Since the torque capacity of the joints are determined by the parallel belt coupling mechanism, the co-design approach is applied in both joint space and actuation space to enable a comparative validation of the proposed method.
The explicit constraint matrix that maps the joint torques to motor torques is given by
\begin{equation}
	\vect{G} =
	\begin{bmatrix}
	{1}/{g_1} & 0 & 0 & 0  \\
	{1}/{g_2} & {1}/{g_2} & 0 & 0  \\
	1 & {1}/{g_3} & {1}/{g_4} & {1}/{g_4}   \\
	1 & {1}/{g_3} & {1}/{g_4} & - {1}/{g_4} 
	\end{bmatrix}
\end{equation}
where, $g_1 \dots g_4$ represent the gear ratios of the belt couplings. 
Note that the gear ratios are continuous, as they are calculated as the quotient of the radii of the gear wheels to which the belt is attached.
This explicit constraint matrix $\vect{G}$ establishes a direct connection between the joint space and the actuation space, influencing the torque distribution.
In this special case, with constant terms in $\vect{G}$, $\vect{g}_u = \vect{0}$.
By optimizing these gear ratios, the manipulator can handle higher payloads while maintaining its dynamic performance.
In our co-design approach, the motion planning is formulated in both, joint space and actuation space, as described in the following sections. 
\subsection{Motion Optimization in Joint Space}
A conventional approach in co-design methodologies is to plan motion directly in the joint space of a robot, for example by using OC. Here, state variables (position and velocity) and torque control inputs are formulated accordingly. 
However, in the context of a parallel belt mechanism, such an approach does not fully capture the interaction between the actuators and joints. 
Specifically, the differential coupling between Joint 3 and Joint 4 (see Fig.~\ref{fig:topological-graph}) is lost when using a joint space representation. 
Consequently, the torque limits must be properly accounted for using the explicit constraint matrix $\vect{G}$. 
In this formulation, the joint torques and velocity limits are expressed as:
\begin{equation}
 \def\arraystretch{1.2}
\begin{matrix}
\label{eq:joint-space-limits}
\vect{G}^{-1} \dot{\vect{q}}_{u_{\min}} &\leq & \dot{\vect{q}} &\leq& \vect{G}^{-1} \dot{\vect{q}}_{u_{\max}} \\
\vect{G}^T \vect{\tau}_{u_{\min}} &\leq & \vect{\tau} &\leq &\vect{G}^T \vect{\tau}_{u_{\max}}
\end{matrix}
\end{equation}
where $\dot{\vect{q}}_{u}$ and $\vect{\tau}_u$ are the actuator velocity and torque limits respectively.

The motion planning is then formulated as an optimization problem described in (\ref{eq:OCP}). 
Here, velocity and torque limits are handled as described in (\ref{eq:joint-space-limits}).
In this formulation, the dynamic constraints are defined in the joint space without explicitly considering gear ratio coupling in the EoM, as described in (\ref{eq:eom_full_implicit}).

\subsection{Motion Optimization in Actuation Space}
Unlike the conventional approach, where control limits and EoM are described in joint space, we formulate the OC problem in actuation space. This formulation provides a more accurate representation of the parallel coupling dynamics between actuators and joints while ensuring compliance with actuator limits.
Accordingly, the motion optimization problem in actuation space is defined as in (\ref{eq:OCP}), with control limits expressed as:
\begin{equation}
\vect{\tau}_{u_{\min}} \leq \vect{u} \leq \vect{\tau}_{u_{\max}}
\end{equation}
The velocity limits remain unchanged from (\ref{eq:joint-space-limits}), as the state representation still considers joint-space velocity constraints. 
Additionally, the optimization problem is subject to the dynamic constraints specified in (\ref{eq:explicit-eom}).
\subsection{Design Optimization for Payload Adaptability}
To accommodate varying payloads, we optimize the continuous gear ratios $g_1, g_2, g_3, g_4$ as design parameters in $\vect{G}$. 
The gear ratio bounds are imposed based on the mechanical constraints of the manipulator as follows:
\begin{equation}
\begin{aligned}
& 1 \leq g_1, g_2 \leq 9 \\
& 1 \leq g_3, g_4 \leq 3
\end{aligned}
\end{equation}
with
\begin{equation}
\label{eq:gear-ratio}
g_3 \leq g_2 < g_1 
\end{equation}
By optimizing these gear ratios, a modular manipulator can be developed, allowing for adjustments based on the specific payload requirements. 
This modularity ensures that the robotic arm can handle payloads of different weights, while maintaining agility and efficiency.


\subsection{Co-Design Implementation Details}
\begin{figure}[!t]
	\centering
	\includegraphics[width=0.99\linewidth]{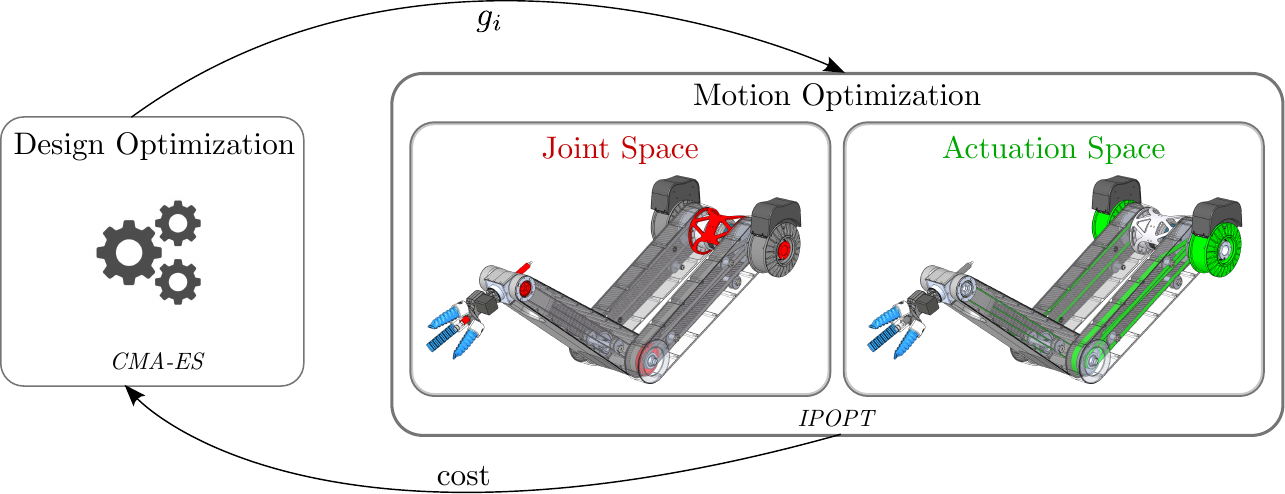}
	\caption{Co-design pipeline with joint space (conventional) and actuation space (ours)} 
	\label{fig:co-design-pipeline}
	\vspace{-0.3cm}
\end{figure}

Our co-design approach is formulated in both, joint space and actuation space, optimizing the mechanical design and control strategies simultaneously in a bi-level formulation. 
The OC problem as described in (\ref{eq:OCP}) is employed to minimize the motion cost while ensuring compliance with the dynamic constraints at the inner level. The outer-level optimizes the gear ratios of the parallel belt couplings. Fig. \ref{fig:co-design-pipeline} illustrates this pipeline. In the figure, the four joints of the manipulator are highlighted in red, while the actuators including the belt couplings are indicated in green.
The inner-level OC problem is formulated using CasADi \cite{casadi}, and the dynamics are computed using the open-source library Pinocchio \cite{pinocchioweb}.
The total duration for trajectory generation is set to $T = 0.7$ s with $N = 50$ discretization steps. 
This selection is slightly faster than the trajectory duration of $T = 0.8$ s reported in \cite{christoph-roland}, with the intent of achieving more dynamic motions while ensuring robust pick-and-place operations for various payloads. 
The hyperparameters of the OC are summarized in Table \ref{tab:motion-parameters}. 
These hyperparameters are determined based on empirical evaluations to ensure a balance between trajectory smoothness and dynamic feasibility in joint space and actuation space.
\begin{table}[!htpb]
	\centering
	\setlength{\tabcolsep}{5pt}
	\caption{Hyperparameters for motion optimization}
	\spaceskip=8pt
	\scriptsize
	\vspace{-0.1cm}
	\begin{tabular}{c c}
		\toprule
		{Hyperparameters} & Weights         \\ \midrule
		State Weights ($\vect{Q}$)          & $\text{diag}(10^{-2}, \dots, 10^{-2})$   \\ \midrule
		Control Weights ($\vect{R}$)        & $\text{diag}(10^{-3}, 10^{-3}, 10^{-3}, 10^{-3})$   \\ \midrule
		Cartesian Boundary Weights ($\vect{\rho}$) & $\text{diag}(10^3, 10^3)$    \\ \bottomrule
	\end{tabular}
	\label{tab:motion-parameters}
\end{table}
At the outer level, we employ CMA-ES from the \textit{pagmo} library \cite{pygmo} for optimizing the gear ratios. 
The optimization is performed with a population size of 100 over 30 generations, with an initial step size of $\sigma = 0.3$. 
Infeasible solutions, where the motion planner fails to generate a valid trajectory, or the constraint in (\ref{eq:gear-ratio}) is violated, are penalized with a high cost ($10^6$) to ensure convergence towards physically feasible gear ratios.

\section{Results and Discussion}
\label{sec:results} 
In this section, we present the results of the evaluation of the proposed co-design approach and validate them through a comprehensive comparison.
First, we use the original manipulator design and analyze the optimized motion in joint space and actuation space to illustrate the inherent differences between the two representations.
Next, we determine the optimal gear ratios using co-design in joint space and actuation space for a payload of 1 kg and 3 kg and compare the resulting optimal manipulator trajectories.
The results are also summarized in the accompanying video\footnote{\url{https://youtu.be/9Izb75ocapk}}.
\subsection{Motion Optimization in Joint Space and Actuation Space}
\begin{figure}[!htpb]
	\centering
	\includegraphics[width = 0.8\linewidth]{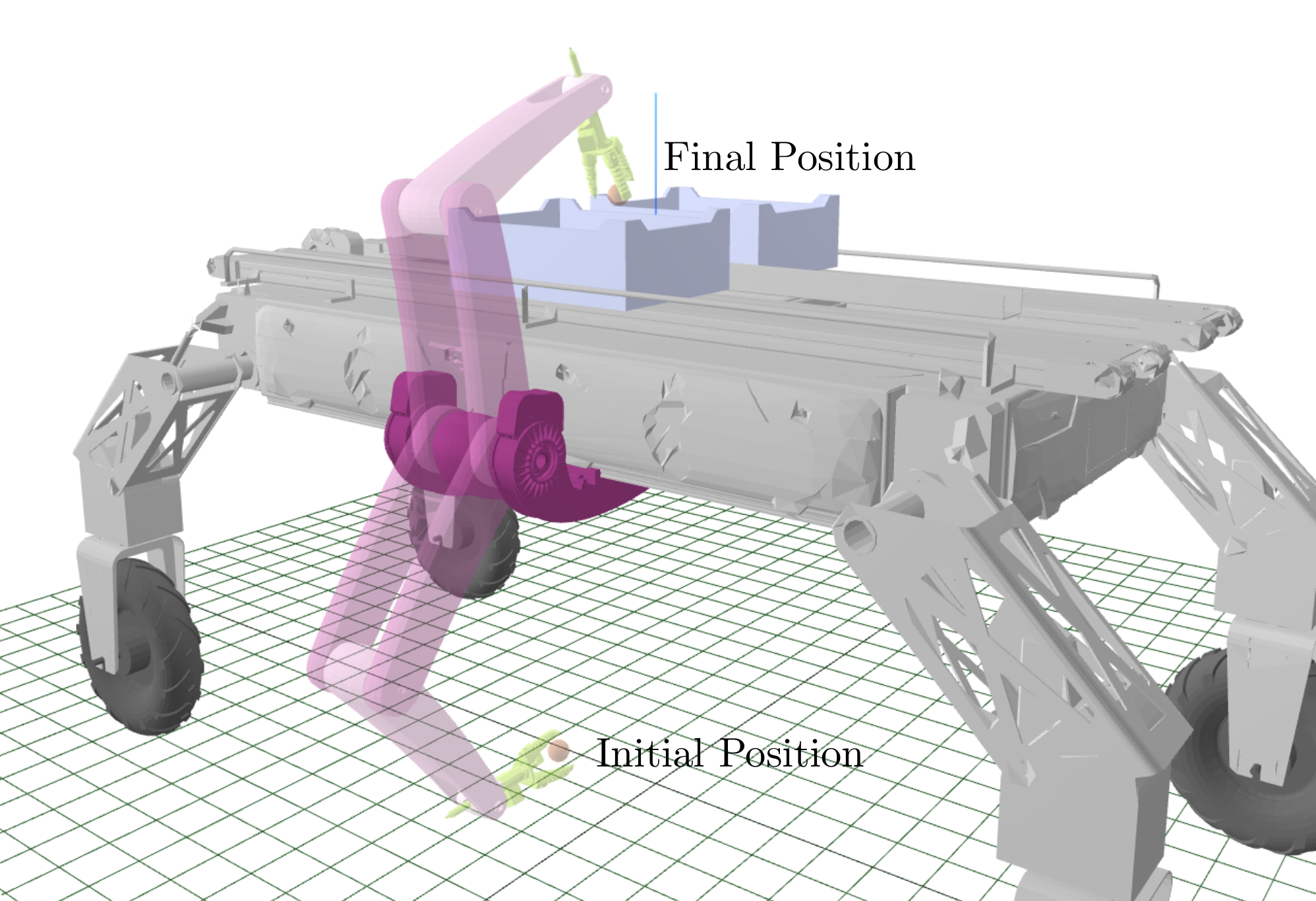}
	\caption{Initial and final position of the manipulator}
	\label{fig:init-final-pos}
	\vspace{-0.3cm}
\end{figure}

\begin{figure}[!htpb]
	\centering
	\begin{minipage}{0.6\linewidth}
		\centering
		\subfigure[Joint positions]{\includegraphics[width = 1.0\linewidth]{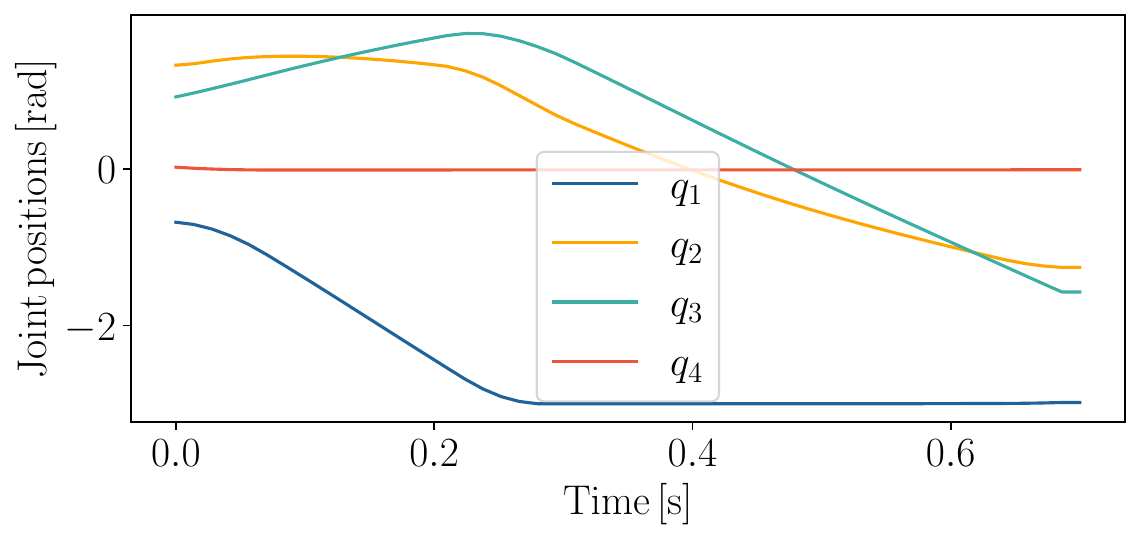}\label{fig:orig_pos_act}} \\
		\subfigure[Joint torques]{\includegraphics[width = 1.0\linewidth]{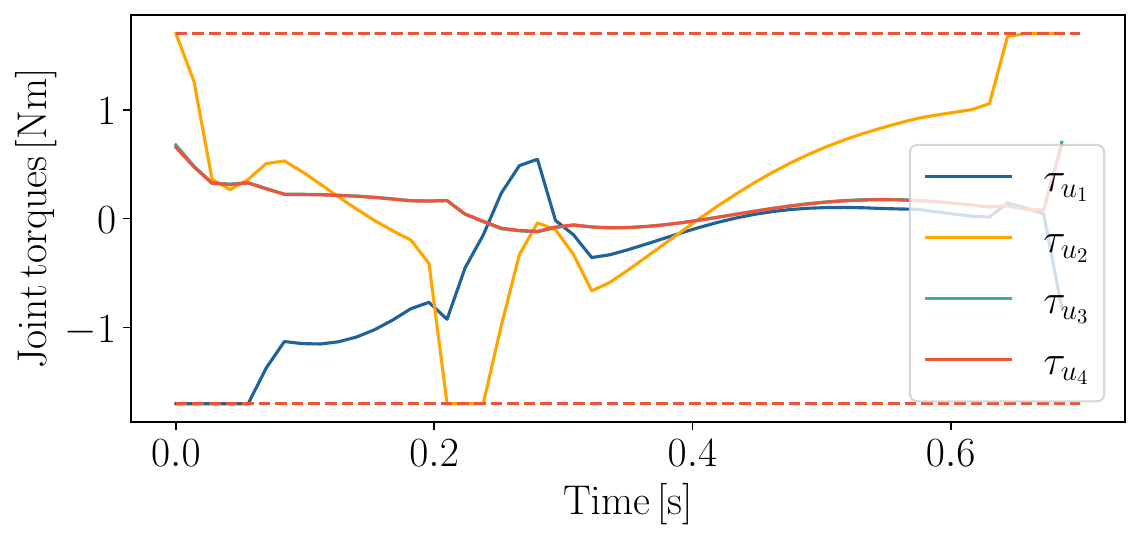}\label{fig:orig_tau_act}}
	\end{minipage}%
	\begin{minipage}{0.40\linewidth}
		\centering
		\subfigure[Cartesian plot]{\includegraphics[width = 1.0\linewidth]{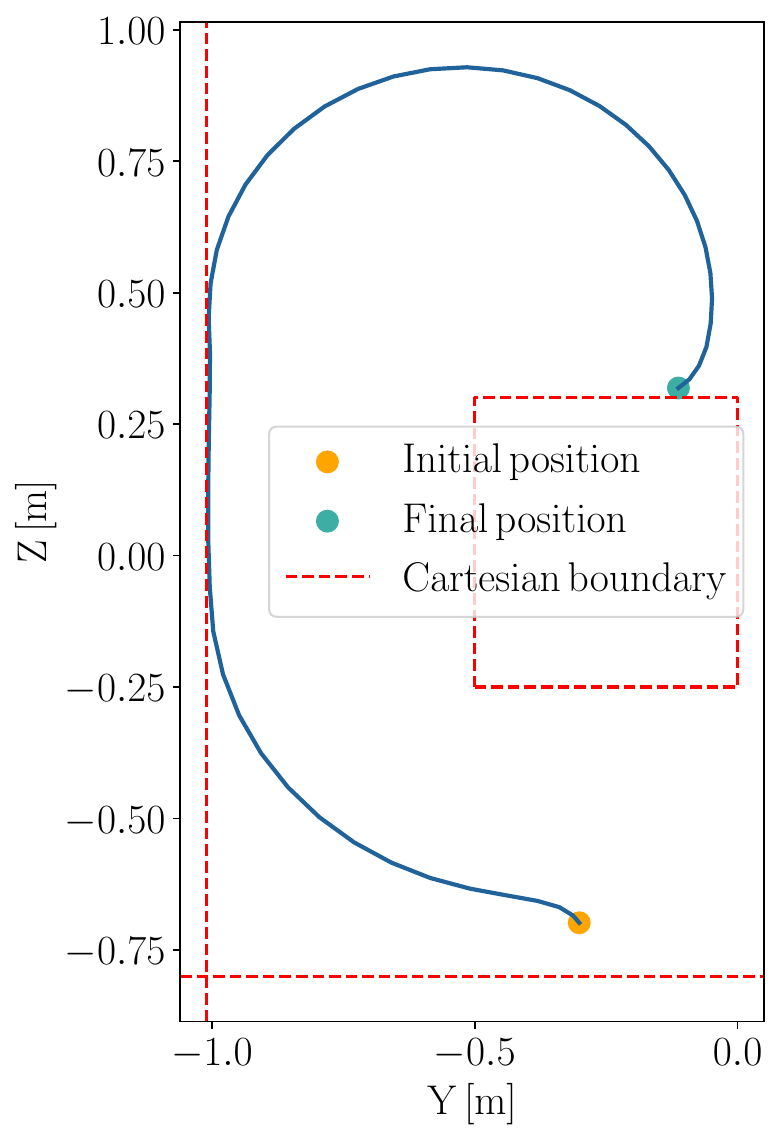}\label{fig:orig_cart_act}}
	\end{minipage}
	\caption{Actuation space motion optimization with original design}
	\label{fig:act-space-orig}
	\vspace{-0.3cm}
\end{figure}

We use the OC problem from~(\ref{eq:OCP}) and the hyperparameters from Table~\ref{tab:motion-parameters} to generate the motion shown in Fig.~\ref{fig:init-final-pos}. Since the manipulator is to be used in  fruit harvesting, the initial position corresponds to the point at which the fruit is picked and the final position corresponds to the point at which the fruit is placed in a collection container.
The OC problem ensures that the planned trajectories satisfy the EoM of the system while minimizing the given cost function. 
The gear ratios in the original design are $[g_1, g_2, g_3, g_4] = [6,3,1,1]$, with a torque limit of $\pm1.7$ Nm for all four motors.
Fig. \ref{fig:act-space-orig} illustrates the resulting joint positions, motor torques and Cartesian space trajectory when the OC problem is formulated in actuation space. 
The resulting motion is represented by a smooth trajectory between the initial and final positions in Cartesian space, while respecting the torque constraints. The differential coupling effect is evident in $\tau_{u,3}$ and $\tau_{u,4}$, which follow identical torque profiles throughout the motion. A slight saturation can be observed in the torque values, but this is acceptable given the maximum torques of the motors.

\begin{figure}[!htpb]
	\centering
	\begin{minipage}{0.6\linewidth}
		\centering
		\subfigure[Joint positions]{\includegraphics[width = 1.0\linewidth]{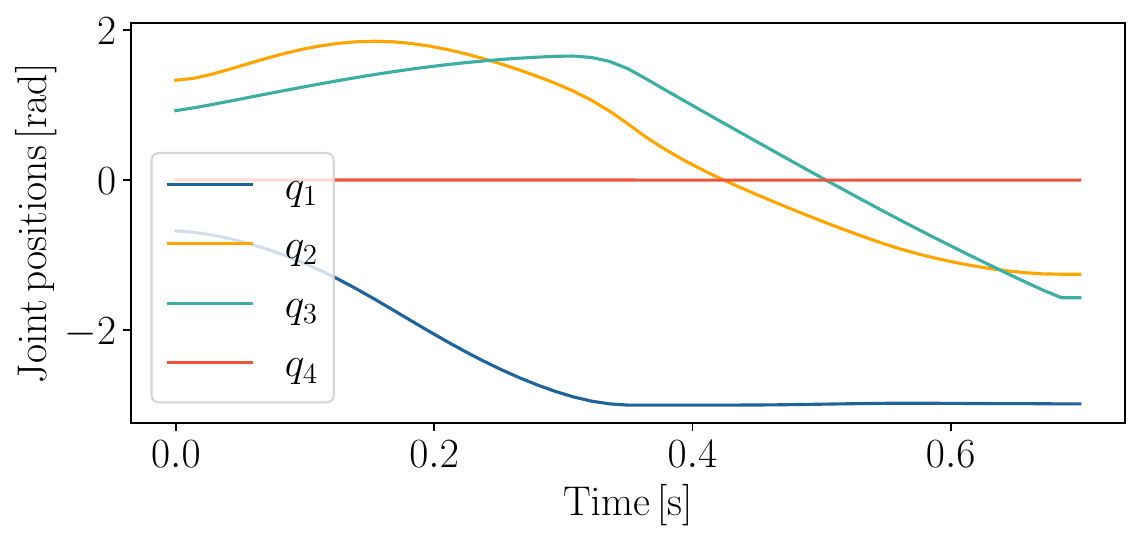}\label{fig:orig_pos_ind}} \\
		\subfigure[Joint torques]{\includegraphics[width = 1.0\linewidth]{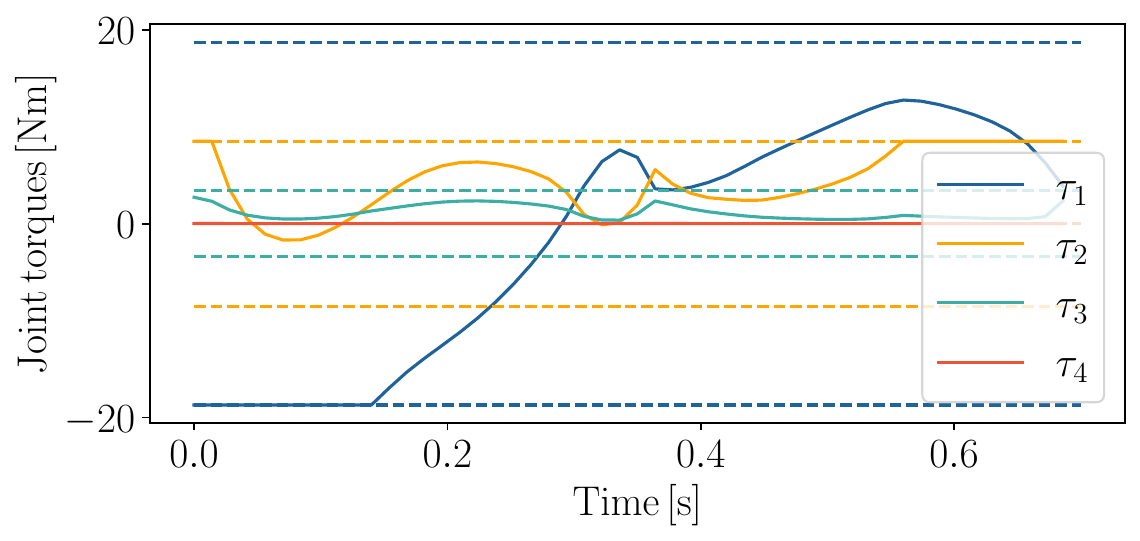}\label{fig:orig_tau_ind}}
	\end{minipage}%
	\begin{minipage}{0.40\linewidth}
		\centering
		\subfigure[Cartesian plot]{\includegraphics[width = 1.0\linewidth]{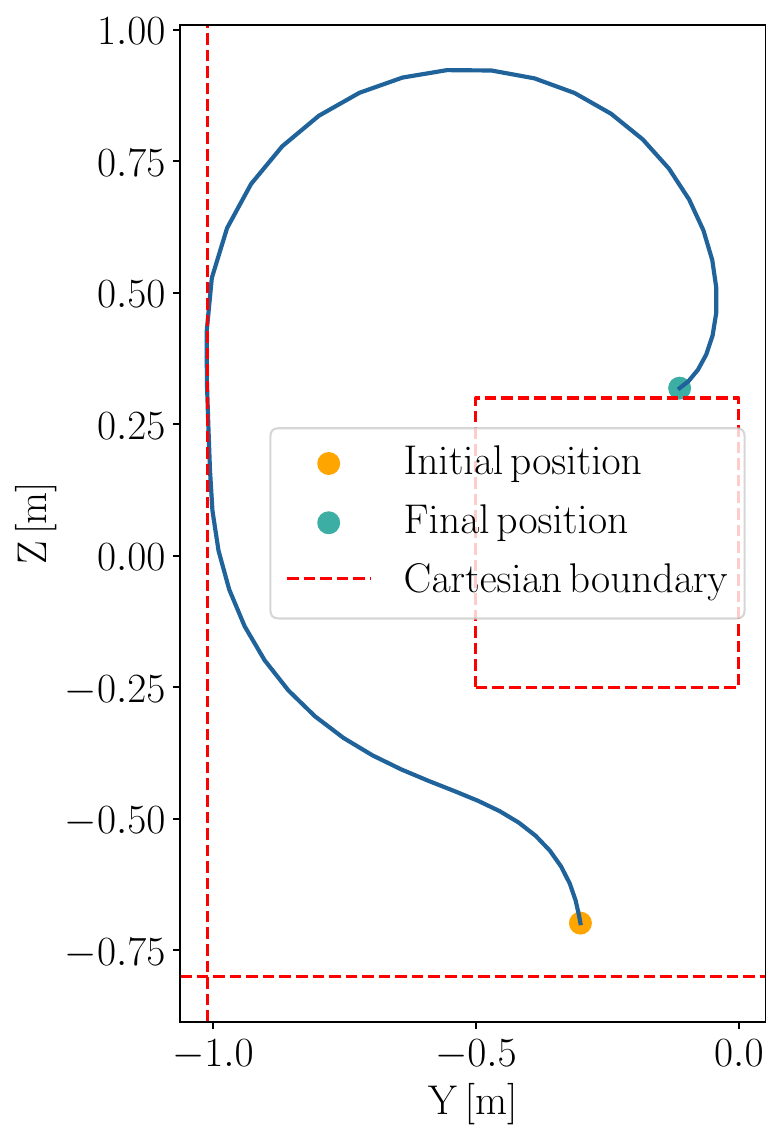}\label{fig:orig_cart_ind}}
	\end{minipage}
	\caption{Joint space motion optimization with original design}
	\label{fig:joint-space-orig}
	\vspace{-0.3cm}
\end{figure}

The results of joint-space motion optimization are shown in Fig. \ref{fig:joint-space-orig}. The torque limits imposed by the parallel coupling on the joints are given by $\vect{\tau}_{max},\vect{\tau}_{min} = [\pm 18.7, \pm 8.5, \pm 3.4, \pm 0]$ Nm. The Cartesian space trajectory looks slightly different when comparing actuation-space and joint-space motion planning. However, both approaches produce feasible solutions.
\subsection{Co-design in Joint Space and Actuation Space}
In this section, we describe the results on performing co-design in joint space and actuation space of the belt-driven robot manipulator. To find optimal gear ratios, we use CMA-ES as a black-box optimizer in the design space. 
To obtain meaningful results, we run the entire co-optimization five times for each case, with different random seeds for the initial population. 
The optimization criterion used in CMA-ES is chosen to be identical to the cost function in~(\ref{eq:OCP}). 
\begin{figure}[!htpb]
	\centering
	\includegraphics[width = 0.95\linewidth]{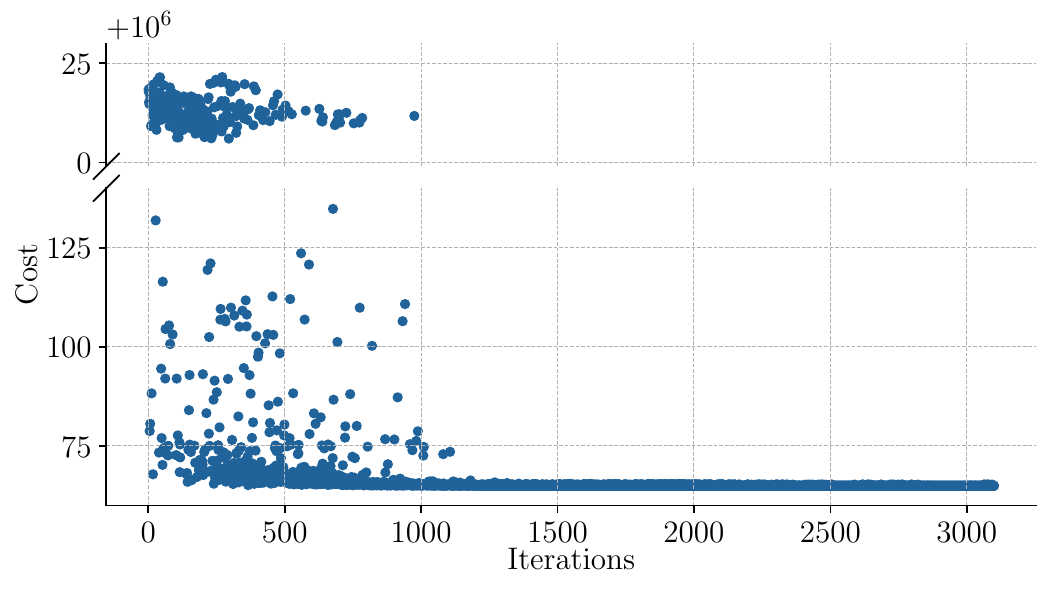}
	\caption{Evolution of the cost function in CMA-ES}
	\label{fig:population-evolution}
	\vspace{-0.3cm}
\end{figure}
\subsubsection{1kg Payload}
\begin{figure}[!t]
	\centering
	\begin{minipage}{0.49\linewidth}
		\centering
		\subfigure[Joint torques (original)]{\includegraphics[width = 1.0\linewidth]{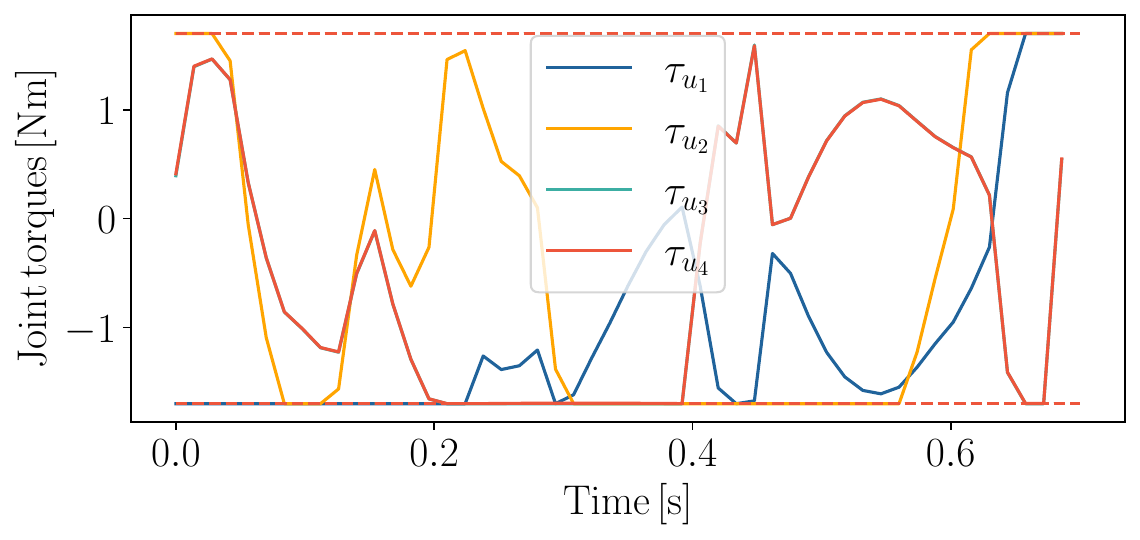}\label{fig:1kg_orig_tau_act}} \\
		\subfigure[Cartesian plot (original)]{\includegraphics[width = 1.0\linewidth]{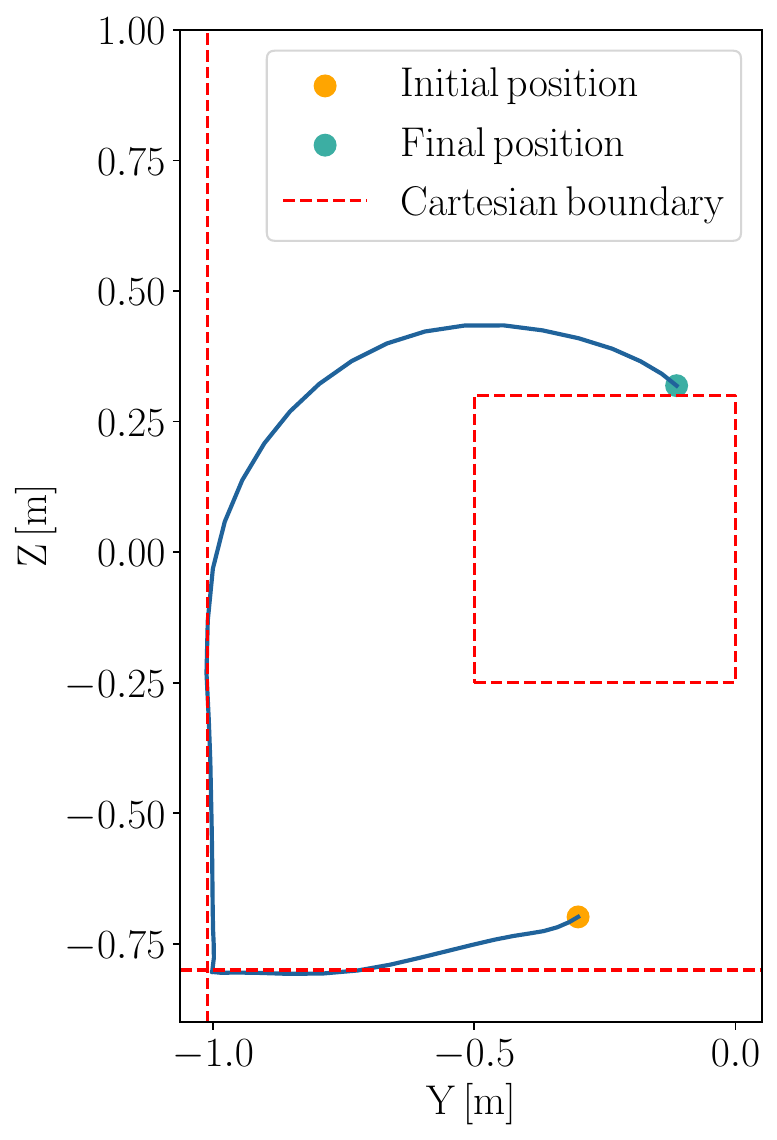}\label{fig:1kg_orig_cart_act}}
	\end{minipage}%
	\begin{minipage}{0.49\linewidth}
		\centering
		\subfigure[Joint torques (optimized)]{\includegraphics[width = 1.0\linewidth]{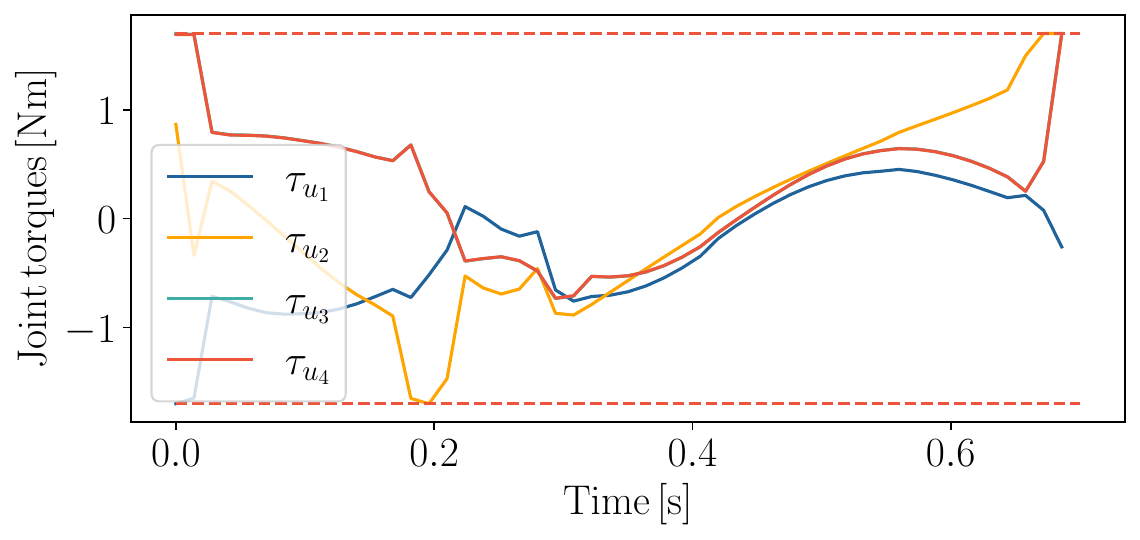}\label{fig:1kg_after_tau_act}} \\
		\subfigure[Cartesian plot (optimized)]{\includegraphics[width = 1.0\linewidth]{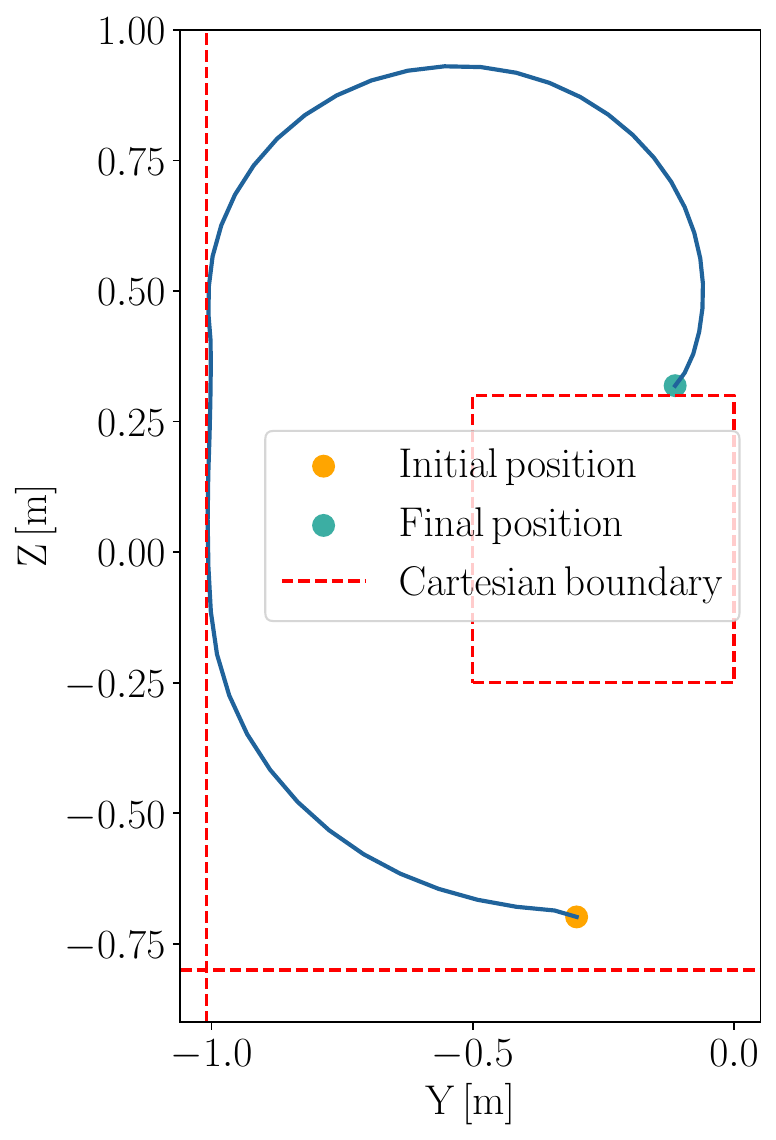}\label{fig:1kg_after_cart_act}}
	\end{minipage}
	\caption{Comparison of joint torques and cartesian trajectories before and after co-design in actuation space for 1kg.}
	\label{fig:act-space-1kg}
	\vspace{-0.3cm}
\end{figure}

In order to evaluate the capabilities of our approach to optimize the load capacity of the manipulator, tests are performed with a payload of 1 kg. 
The evolution of the cost function of CMA-ES is shown in Fig. \ref{fig:population-evolution}. 
In this graph, the populations that generate infeasible solutions can be seen with costs around $10^6$. The graph shows that CMA-ES converges after about 1200 iterations.
Using the original robot design, no feasible solution could be found when formulating the OC problem in joint space. When formulating the OC problem in the actuation space, a feasible solution could still be obtained. 
However, the quality of the solution was low due to torque saturation (see Fig.~\ref{fig:1kg_orig_tau_act}) and a non-smooth Cartesian space trajectory (see Fig.~\ref{fig:1kg_orig_cart_act}). 
After applying our co-design approach, the resulting motion is a smooth trajectory (Fig.~\ref{fig:1kg_after_cart_act}) without saturation of the motor torques (Fig. \ref{fig:1kg_after_tau_act}). 
\begin{figure}[!t]
	\centering
	\begin{minipage}{0.6\linewidth}
		\centering
		\subfigure[Joint rotations]{\includegraphics[width = 1.0\linewidth]{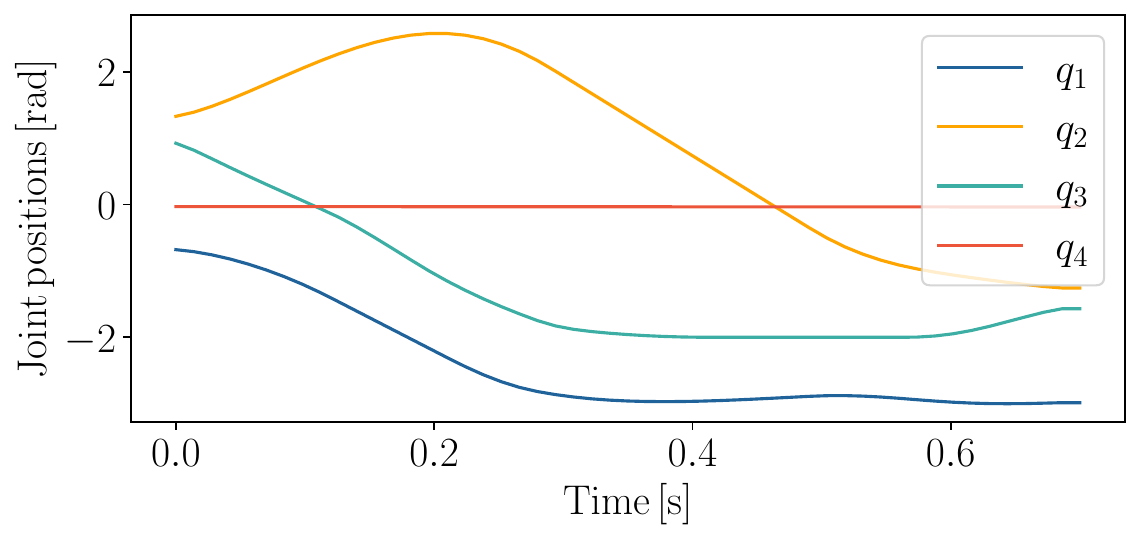}\label{fig:1kg_after_pos_ind}} \\
		\subfigure[Joint torques]{\includegraphics[width = 1.0\linewidth]{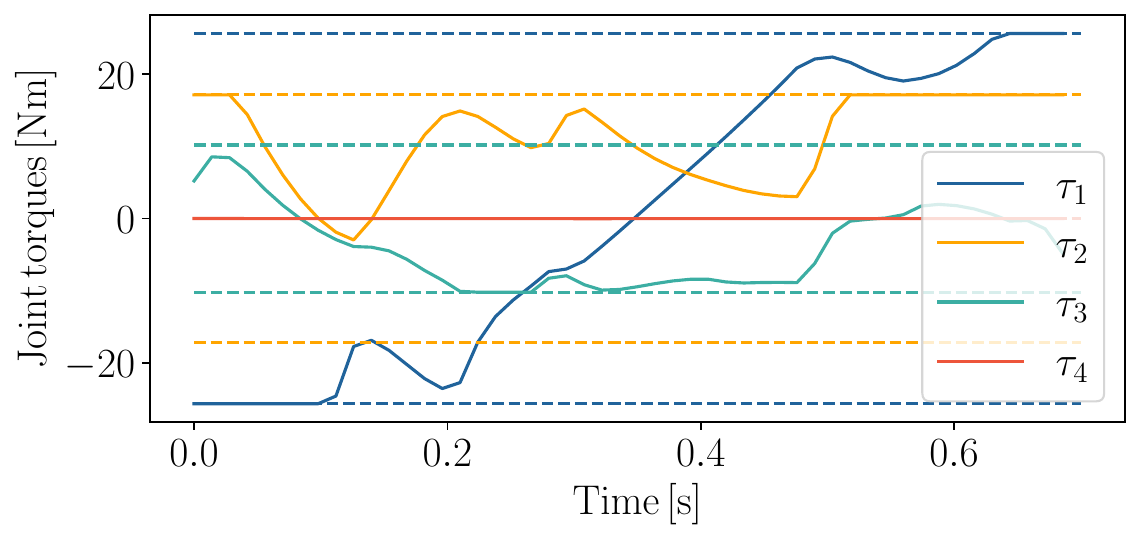}\label{fig:1kg_after_tau_ind}}
	\end{minipage}%
	\begin{minipage}{0.40\linewidth}
		\centering
		\subfigure[Cartesian plot]{\includegraphics[width = 1.0\linewidth]{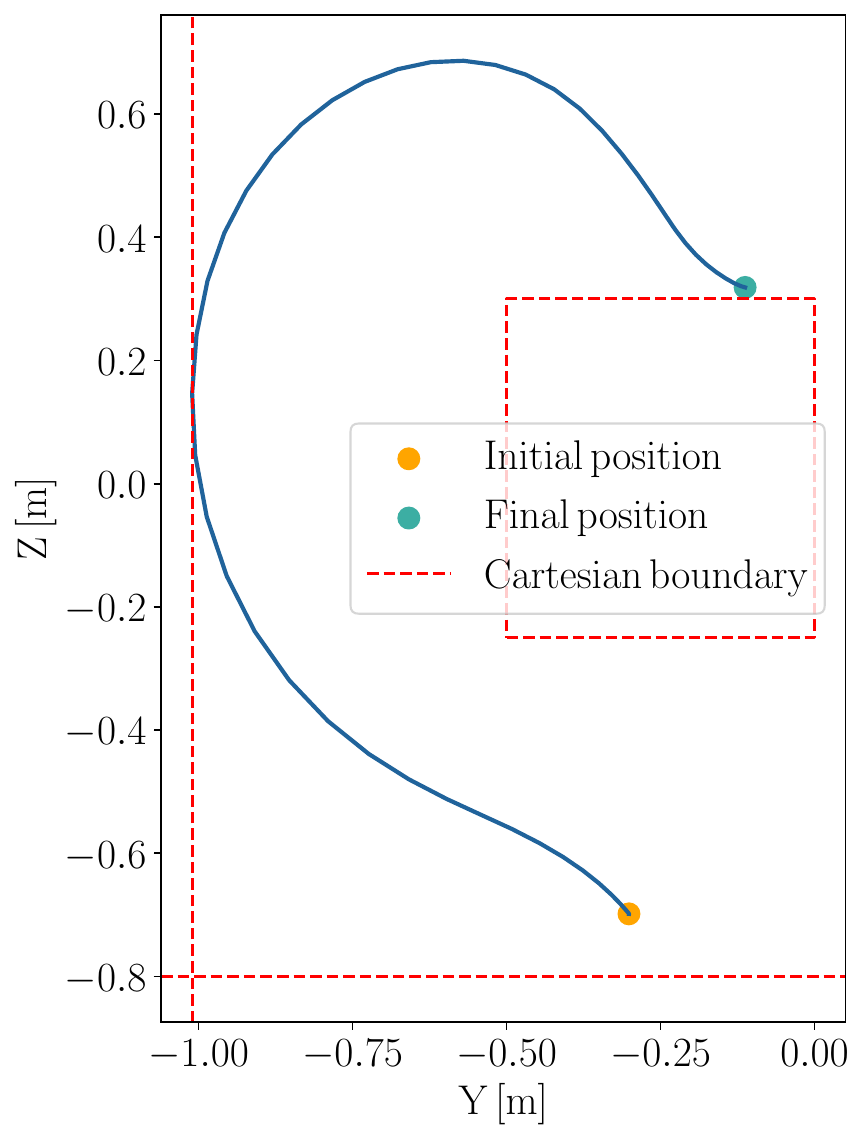}\label{fig:1kg_after_cart_ind}}
	\end{minipage}
	\caption{After co-design in joint space for 1kg with optimal gear ratios}
	\label{fig:ind-space-1kg}
	\vspace{-0.3cm}
\end{figure}
Similarly, when performing co-design in joint space, a feasible solution could be obtained. As shown in Fig. \ref{fig:ind-space-1kg}, the resulting Cartesian trajectory is smooth and adheres to the imposed constraints. 
However, in contrast to the solution obtained by actuation space co-design, the joint torques consistently reach saturation.

\subsubsection{3kg Payload}
\begin{figure}[!htpb]
	\centering
	\begin{minipage}{0.49\linewidth}
		\centering
		\subfigure[Joint torques (original)]{\includegraphics[width = 1.0\linewidth]{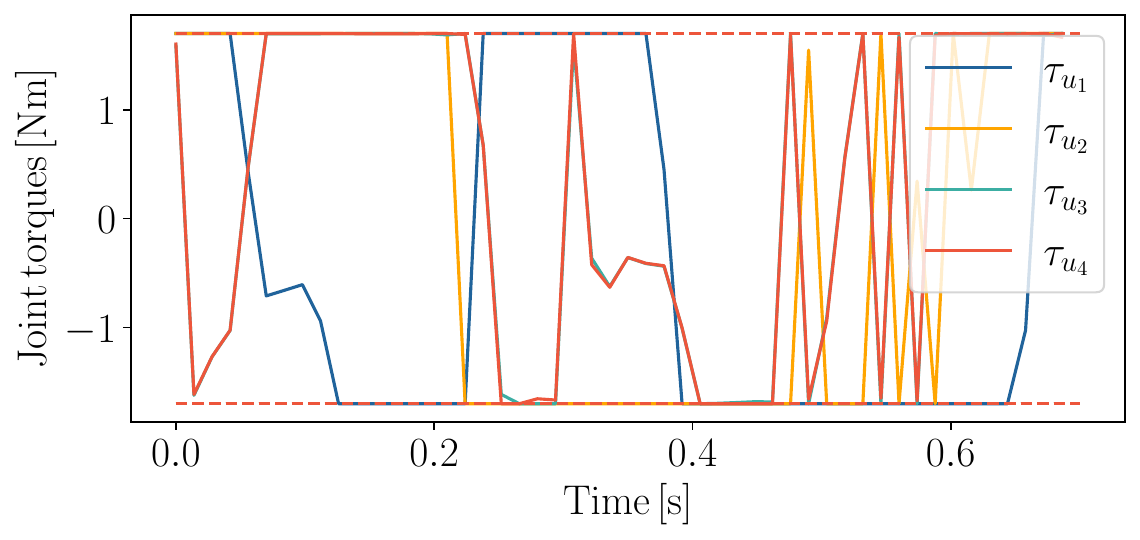}\label{fig:3kg_orig_tau_act}} \\
		\subfigure[Cartesian plot (original)]{\includegraphics[width = 1.0\linewidth]{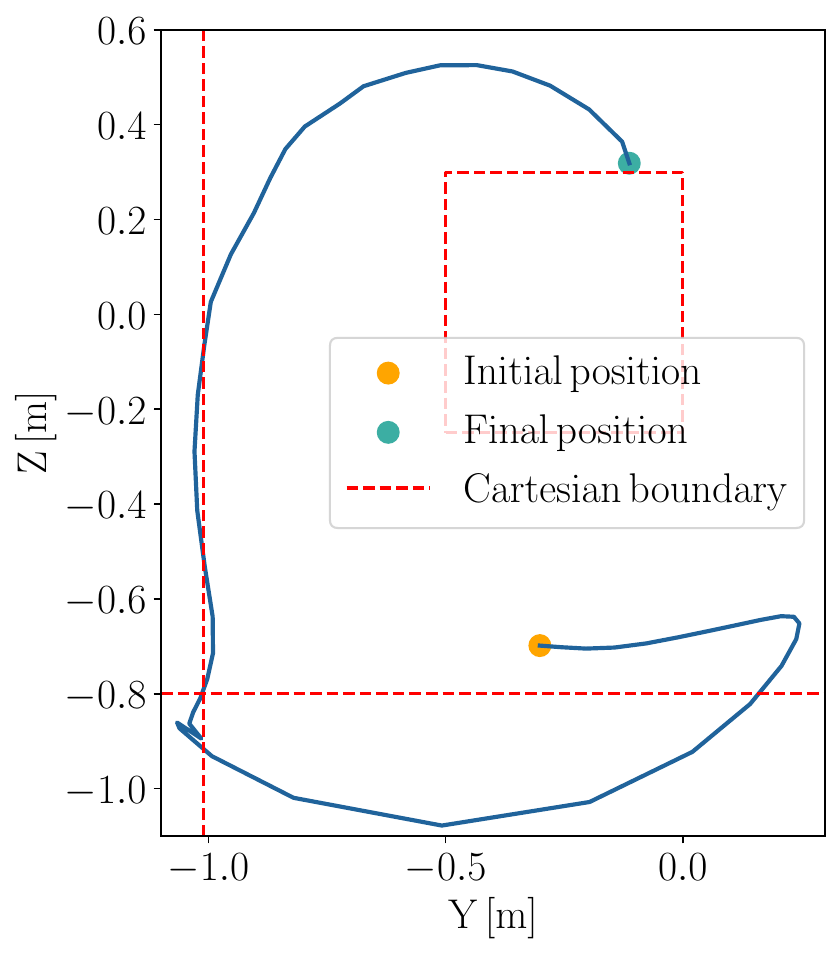}\label{fig:3kg_orig_cart_act}}
	\end{minipage}%
	\begin{minipage}{0.49\linewidth}
		\centering
		\subfigure[Joint torques (optimized)]{\includegraphics[width = 1.0\linewidth]{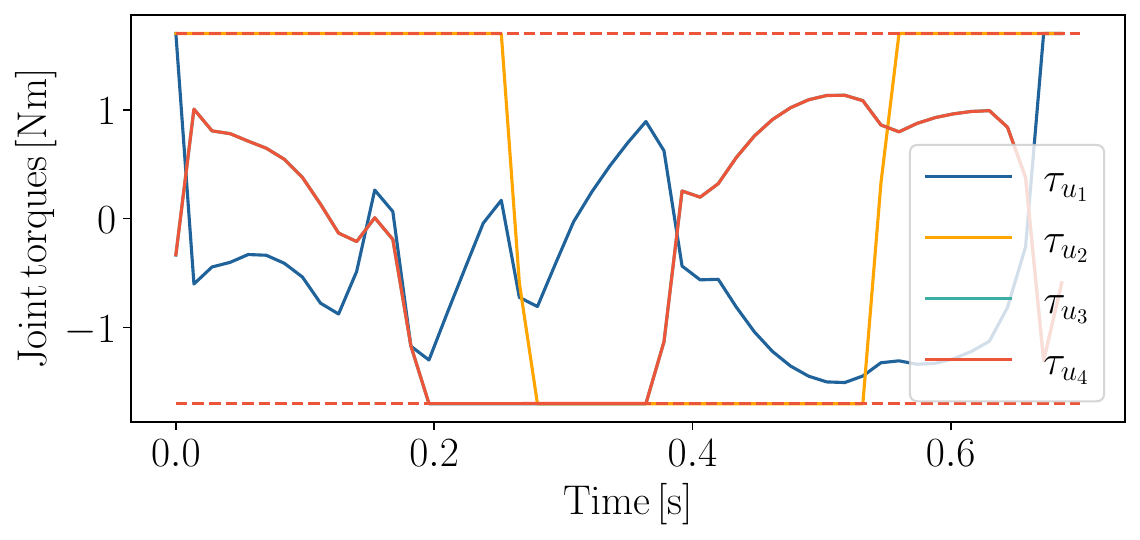}\label{fig:3kg_after_tau_act}} \\
		\subfigure[Cartesian plot (optimized)]{\includegraphics[width = 1.0\linewidth]{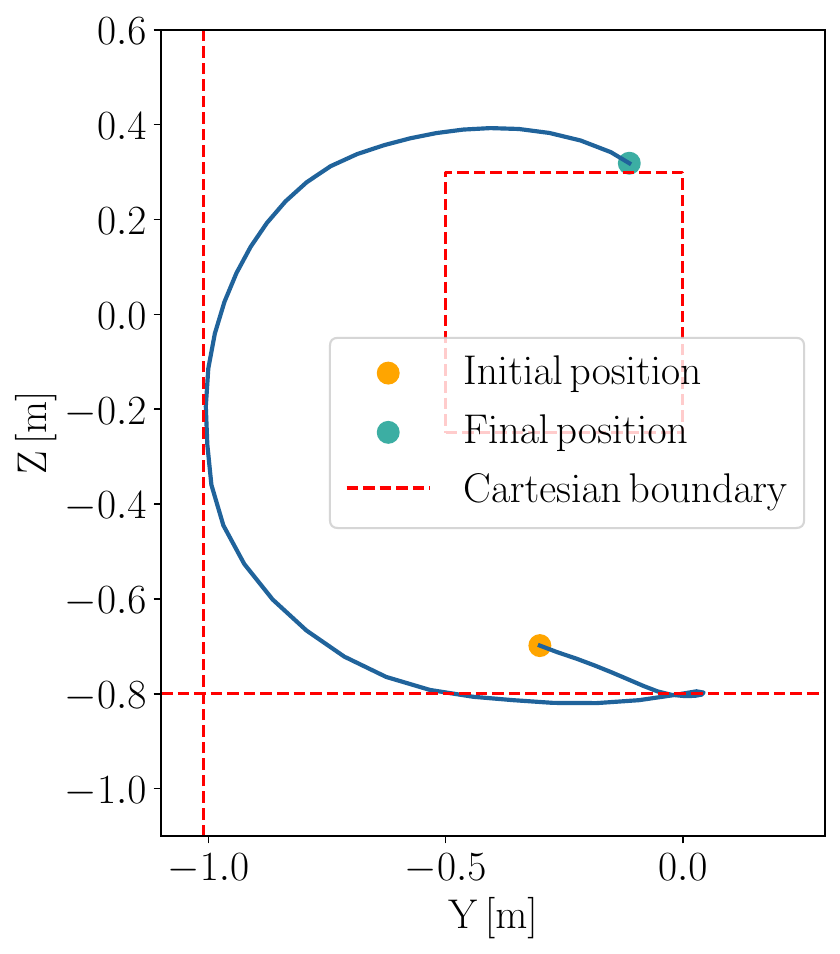}\label{fig:3kg_after_cart_act}}
	\end{minipage}
	\caption{Comparison of joint torques and cartesian trajectories before and after co-design in actuation space for 3kg.}
	\label{fig:act-space-3kg}
	\vspace{-0.3cm}
\end{figure}

To determine the limitations of the co-design approach presented, further experiments were conducted with a payload of 3 kg. With the original design, the manipulator design was not able to handle this load effectively. 
As shown in Fig. \ref{fig:3kg_orig_cart_act}, the manipulator exhibits an initial pendulum motion before attempting to lift the payload. 
The trajectory strongly violates the Cartesian space constraints.
In contrast, after performing the co-design in the actuation space, the resulting motion is smooth and adheres to the imposed Cartesian position constraints. 
Fig.~\ref{fig:3kg_after_cart_act} shows that the optimized manipulator initially performs a swing-up of the payload, while adhering to the Cartesian space boundaries, before successfully lifting the load and placing it at the final position.

In contrast, the co-design approach in the joint space is not capable to produce a manipulator design that can handle the desired payload of 3 kg in this task.


\subsection{Discussion}

\begin{figure}[!htpb]
	\centering	
	\subfigure[Joint Space]{\includegraphics[width = 0.49\linewidth]{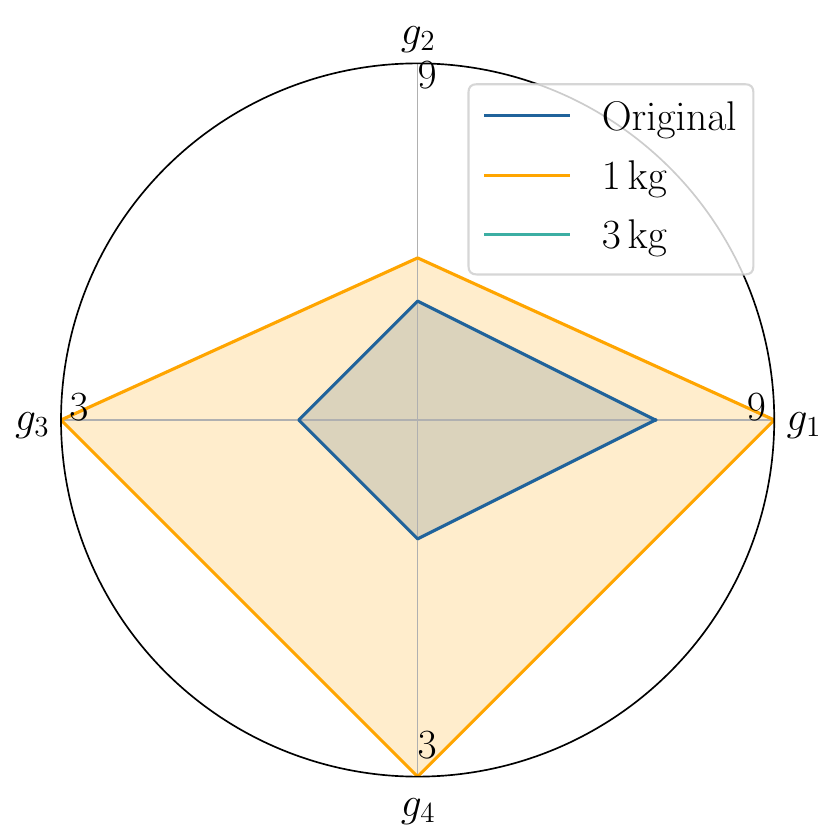}\label{fig:spider_ind}}
	\subfigure[Actuation Space]{\includegraphics[width = 0.49\linewidth]{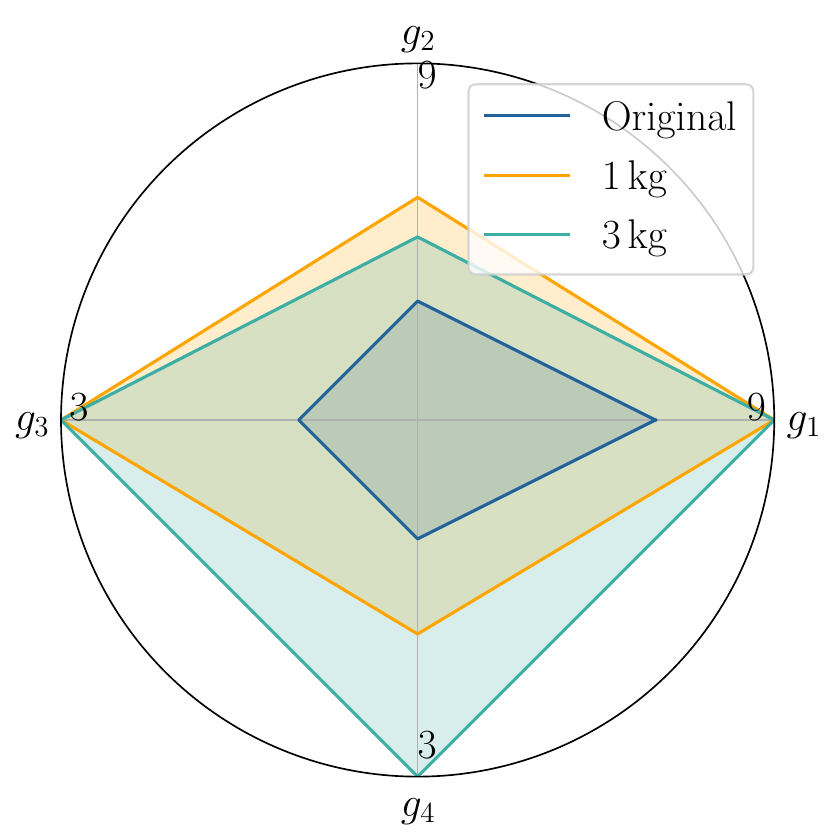}\label{fig:spider_act}}
	\caption{Optimal gear ratios for different payloads}
	\label{fig:spider-plots}
	\vspace{-0.3cm}
\end{figure}

The previous section presented results on comparing joint space and actuation space co-design of a belt-driven robotic manipulator.
The co-design results in joint space and actuation space are summarized in Table \ref{tab:results-table}, where before and after corresponds to original design and optimal design respectively. 
The optimal gear ratios obtained are also illustrated in Fig. \ref{fig:spider-plots}. 
When using joint space co-design, the resulting gear ratios $g_3$ and $g_4$ already reach their limits with a payload of 1 kg. 
In contrast, when using actuation space co-design, the resulting gear ratios are much smaller for a 1 kg payload, while they reach their upper bounds when the payload increases to 3 kg.
Our co-design implementation, applied to a robot with moving mass of 1.6 kg, doubled the manipulator's payload capability, approximately to 3 kg. This improvement was achieved within the system's maximum gear ratio limits, requiring only gear adaptations while keeping the motors unchanged. 
The inclusion of parallel coupling constraints in our approach enabled significant capability enhancements with minimal cost and weight addition.
The results indicate clearly that co-design in actuation space is more efficient, as it better exploits the available solution space. 
\begin{table}[!htpb]
	\centering
	\setlength{\tabcolsep}{5pt}
	\caption{Gear Ratios and Cost for Different Payloads}
	\spaceskip=8pt
	\scriptsize
	\vspace{-0.1cm}
	\begin{tabular}{c c c c c}
		\toprule
		Payload & Space & Before/After & Gear Ratios & Cost \\ 
		\midrule
		\multirow{2}{*}{No payload} & {Joint} & \multirow{2}{*}{Before}  & \multirow{2}{*}{$[6,3,1,1]$} & 66.77 \\
		& {Actuation} &    						&                           & 50.99 \\ 
		\midrule							
		\multirow{4}{*}{1kg} & \multirow{2}{*}{Joint} & Before  & $[6,3,1,1]$ & - \\ 
		&                         & After   & $[9, 4.09, 3, 3]$ & 146.42 \\ 
		\cmidrule(lr){2-5}
		& \multirow{2}{*}{Actuation} & Before  & $[6,3,1,1]$ & 161.21 \\  
		&                            & After   & $[9, 5.62, 3, 1.8]$ & 71.55 \\ 
		\midrule
		\multirow{4}{*}{3kg} & \multirow{2}{*}{Joint} & Before  & - & - \\ 
		&                         & After   & - & - \\ 
		\cmidrule(lr){2-5}
		& \multirow{2}{*}{Actuation} & Before  & $[6,3,1,1]$ & 526.45 \\  
		&                            & After   & $[9, 4.62, 3, 3]$ & 159.42 \\ 
		\bottomrule
	\end{tabular}
	\label{tab:results-table}
\end{table}

However, the effect of adding or removing mass when changing the gear ratios is neglected in our study. 
Naturally, the use of belt transmission introduces several practical challenges, such as backlash, efficiency losses, and friction, which are not considered in this study. 
The primary objective is to maintain a minimalistic co-design strategy that optimizes the use of actuation space to increase the payload  of the manipulator.

While one could consider advanced motion optimization algorithms such as iterative Linear Quadratic Regulator (iLQR) or Differential Dynamic Programming (DDP), we formulate motion optimization as NLP problem, as it allows us to impose both, hard and soft constraints. 
Specifically, initial and final positions, system dynamics, and actuation limits are treated as hard constraints, while the Cartesian boundary conditions, state regularization, and control regularization are incorporated into the cost function of the optimization problem.
For the design optimization process, we employed CMA-ES as a gradient-free black box optimizer.

\section{Conclusion and Outlook}
\label{sec:conclusion} 
In this work, a novel co-design approach that integrates parallel coupling constraints into the dynamic model of a robot is presented, particularly focusing on a belt-driven manipulator. Through a bi-level optimization process, we simultaneously optimize the robot's design, specifically the gear ratios, and its behavior for specific tasks to lift heavy payloads. By optimizing in the actuation space, our approach allows for better exploitation of the manipulator's dynamic range, leading to a significant increase in payload compared to conventional co-design methods based on simplified tree-type models.

Future research will address practical challenges like backlash, efficiency losses, and friction from belts and gears, which were not  considered in this study. 
Including these factors in the optimization process could enhance the performance of the real system and the practical applicability of our approach. 
Furthermore, the co-design strategy introduced in this work will be applied to more complex robots like humanoids to enhance their stability and agility.


%

\bibliographystyle{IEEEtran}
\bibliography{references}

\end{document}